\documentclass[conference]{IEEEtran}
\IEEEoverridecommandlockouts
\usepackage{cite}
\usepackage{amsmath,amssymb,amsfonts}
\usepackage{graphicx}
\usepackage{textcomp}
\usepackage{xcolor}
\usepackage{graphicx}
\usepackage{tabularx}
\usepackage{booktabs}
\usepackage{algorithm}
\usepackage{algpseudocode}
\usepackage{caption}
\usepackage{subcaption}
\usepackage[letterpaper,top=0.75in,bottom=1.07in,left=0.625in,right=0.625in]{geometry}
\usepackage{float}  
\def\BibTeX{{\rm B\kern-.05em{\sc i\kern-.025em b}\kern-.08em
    T\kern-.1667em\lower.7ex\hbox{E}\kern-.125emX}}
\begin{document}

\title{Tri-Select: A Multi-Stage Visual Data Selection Framework for Mobile Visual Crowdsensing
\thanks{* Corresponding author.}
}

\author{
\IEEEauthorblockN{Jiayu Zhang}
\IEEEauthorblockA{\textit{School of Software} \\
\textit{Northwestern Polytechnical University}\\
Xi’an, China 710029 \\
jiayuzhang@mail.nwpu.edu.cn}
\and
\IEEEauthorblockN{Kaixing Zhao$^{\ast}$}
\IEEEauthorblockA{\textit{School of Software} \\
\textit{Northwestern Polytechnical University}\\
Xi’an, China 710029 \\
kaixing.zhao@nwpu.edu.cn}
\and
\IEEEauthorblockN{Tianhao Shao}
\IEEEauthorblockA{\textit{School of Software} \\
\textit{Northwestern Polytechnical University}\\
Xi’an, China 710029 \\
tianhaoshao@mail.nwpu.edu.cn}

\and
\IEEEauthorblockN{Bin Guo}
\IEEEauthorblockA{\textit{School of Computer Science} \\
\textit{Northwestern Polytechnical University}\\
Xi’an, China 710029 \\
guob@nwpu.edu.cn}
\and
\IEEEauthorblockN{Liang He}
\IEEEauthorblockA{\textit{School of Software} \\
\textit{Northwestern Polytechnical University}\\
Xi’an, China 710029 \\
2021050018@nwpu.edu.cn}
}

\maketitle

\begin{abstract}
Mobile visual crowdsensing enables large-scale, fine-grained environmental monitoring through the collection of images from distributed mobile devices. However, the resulting data is often redundant and heterogeneous due to overlapping acquisition perspectives, varying resolutions, and diverse user behaviors. To address these challenges, this paper proposes Tri-Select, a multi-stage visual data selection framework that efficiently filters redundant and low-quality images. Tri-Select operates in three stages: (1) metadata-based filtering to discard irrelevant samples; (2) spatial similarity-based spectral clustering to organize candidate images; and (3) a visual-feature-guided selection based on maximum independent set search to retain high-quality, representative images. Experiments on real-world and public datasets demonstrate that Tri-Select improves both selection efficiency and dataset quality, making it well-suited for scalable crowdsensing applications.
\end{abstract}


\section{Introduction}\label{chap3_1}

Mobile visual crowdsensing (MVC) has emerged as a promising paradigm that harnesses the sensing capabilities of distributed mobile devices—such as smartphones, dashcams, and drones—to collect visual data for large-scale environmental perception and analysis~\cite{guo2017emergence}. By exploiting the ubiquity of camera-equipped devices and the mobility of users, MVC enables dynamic, fine-grained, and real-time monitoring in various application domains, including traffic surveillance, disaster response, urban planning, and environmental protection~\cite{costa2019cityspeed,dautaras2021mobile}.

Despite its widespread potential, MVC systems often face critical challenges arising from the uncontrolled nature of data acquisition. In particular, the visual data collected is typically \textit{redundant, heterogeneous, and unstructured}~\cite{zhang2021secure, marjanovic2018autonomous}. Redundancy is caused by multiple users capturing images from similar or overlapping viewpoints, often repeatedly and without coordination~\cite{hamrouni2019photo}. Heterogeneity stems from differences in camera quality, resolution, shooting angles, illumination, and user behavior~\cite{zhu2018spatiotemporal}. These factors collectively lead to excessive transmission overhead, increased storage demand, and inefficiencies in downstream tasks such as model training or event detection~\cite{mowafi2022distributed}.

\begin{figure}[!t]
    \centering
    \includegraphics[width=1\columnwidth]{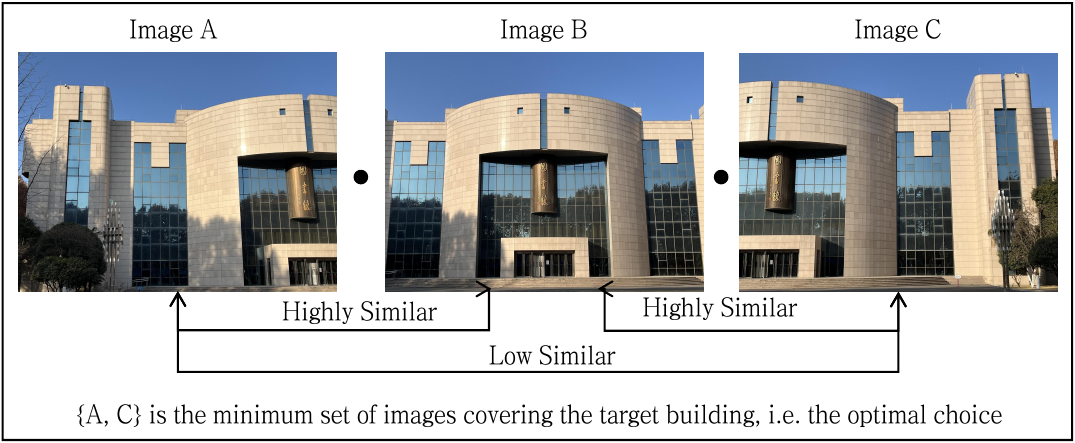}
    \caption{Example of Image Selection in Visual Crowdsensing}
    \label{fig:simPhoto}
\end{figure}

A common deployment scenario involves a swarm of users capturing the same object or region (e.g., a collapsed building or a public event) from different locations and at different times~\cite{hamrouni2019photo,chen2017generic}. As shown in Figure~\ref{fig:simPhoto}, many of the captured images (e.g., image B) may convey visually similar content, adding little incremental value while consuming significant communication and computation resources~\cite{song2022hybrid}. Transmitting and processing the full dataset without intelligent selection is thus both costly and unnecessary. Therefore, there is a pressing need for effective visual data filtering strategies that can reduce redundancy, improve representativeness, and retain essential environmental information~\cite{guo2016worker,chen2017generic}.

To this end, we propose \textbf{Tri-Select}, a lightweight and scalable three-stage visual data selection framework tailored for mobile crowdsensing scenarios. Our goal is to select a minimal set of high-quality, diverse, and task-relevant images from a pool of redundant candidates. Tri-Select is designed with the following key principles in mind: efficiency, representativeness, and modularity~\cite{mathew2022crowdpower,yu2024crowdkit}.

The Tri-Select framework proceeds in three distinct stages: (1) a \textit{metadata-based pre-filtering} phase eliminates low-quality or contextually irrelevant images using spatiotemporal and resolution metadata (e.g., timestamp, GPS coordinates, altitude)~\cite{zhang2020mobile}; (2) a \textit{spatial similarity-based spectral clustering} algorithm organizes the remaining candidates into spatially coherent groups based on acquisition geometry~\cite{chen2017generic}; (3) within each cluster, a \textit{visual feature-guided selection} module applies a maximum independent set (MIS) search over a similarity graph constructed from SIFT descriptors to extract a representative, non-redundant subset of images~\cite{song2022hybrid}.

As illustrated in Figure~\ref{fig:simPhoto}, this multi-stage selection process preserves critical coverage (e.g., images A and C) while eliminating unnecessary duplication (e.g., image B). By jointly exploiting both metadata and image content, Tri-Select ensures scalable and high-quality data selection for downstream visual analytics~\cite{li2023incentive,kalogiros2018allergymap,chen2021blockchain}.

The contributions of this work can be summarized as follows:
\begin{itemize}
    \item We identify and address the critical challenge of visual data redundancy and heterogeneity in mobile crowdsensing applications.
    \item We propose Tri-Select, a novel three-stage framework that integrates metadata analysis, spatial clustering, and visual feature filtering for efficient and effective image selection.
    \item We evaluate the framework on real-world and public datasets, demonstrating its superiority in terms of data reduction, computational efficiency, and coverage quality.
\end{itemize}

The remainder of this paper is organized as follows: Section II reviews related work in visual data selection. Section III presents the problem formulation. Section IV details the proposed Tri-Select framework. Section V evaluates the method using real-world and benchmark datasets. Section VI concludes the paper and discusses future directions.

\section{Related Work}\label{chap3_2}

High-quality data selection is a fundamental problem in visual crowdsensing. Existing research has addressed this challenge from several perspectives, including utility-based selection, redundancy reduction, diversity optimization, and resource-efficient transmission~\cite{guo2016worker, wang2019user, hamrouni2020spatial, deng2023cache, chen2016toward, dao2017managing, zhou2018uncertain, zhou2019utility, wang2011photonet, uddin2012photonet+, wang2014smartphoto, guo2014fliermeet, guo2016picpick, chen2017generic, dao2014managing, zuo2018bandwidth, song2022hybrid}. This section reviews these representative approaches and discusses their limitations in handling heterogeneous, large-scale visual data.

\textbf{Utility-based selection} aims to quantify image value for maximizing task relevance. Zhou et al.~\cite{zhou2018uncertain} integrated GPS metadata and SIFT features to support diverse and similar view evaluation, while Zhou et al.~\cite{zhou2019utility} further proposed a spatial-coverage-aware utility model with greedy optimization. Although effective, such models often assume single-task settings and uniform data sources, limiting their adaptability to complex, heterogeneous scenarios.

\textbf{Redundancy-aware methods} focus on filtering duplicate or similar content from user-contributed data. PhotoNet~\cite{wang2011photonet} and its enhanced version PhotoNet+~\cite{uddin2012photonet+} perform semantic-level redundancy filtering and prioritize diversity. SmartPhoto~\cite{wang2014smartphoto} and FlierMeet~\cite{guo2014fliermeet} leverage spatial-temporal and geometric metadata to identify redundant images in real time. However, these methods lack granularity when dealing with fine-scale variations across diverse devices and viewpoints.

\textbf{Diversity-driven frameworks} such as PicPick~\cite{guo2016picpick} and CrowdPic~\cite{chen2017generic} construct visually diverse subsets under task constraints using hierarchical or adaptive clustering. While effective in maintaining diversity, these approaches typically depend on centralized processing, which hinders real-time or edge deployment.

\textbf{Resource-efficient optimization} seeks to reduce transmission or computation overhead. Dao et al.~\cite{dao2014managing} proposed a metadata-first strategy to minimize upload costs, followed by Zuo et al.~\cite{zuo2018bandwidth} who introduced dynamic feature precision control. More recently, Song~\cite{song2022hybrid} combined contextual and content-aware metrics for selection efficiency. Despite these efforts, most assume homogeneous data formats or lack flexibility for multi-stage selection.

In summary, prior research has contributed valuable techniques for improving visual data selection. However, most approaches: (1) rely on full image uploads and centralized models, limiting scalability; (2) focus on homogeneous data without addressing multi-source heterogeneity; and (3) rarely integrate metadata filtering, spatial organization, and visual analysis in a staged manner. To overcome these gaps, our work proposes a multi-stage selection framework that sequentially applies lightweight filtering, spatial similarity clustering, and visual-feature-based redundancy elimination, enabling scalable and efficient selection from large-scale visual crowdsensing data.

\section{Problem Definition}\label{chap3_3}

Efficient image selection in visual crowdsensing requires formal modeling of tasks and data to address challenges such as content redundancy, heterogeneous device perspectives, and transmission bottlenecks. In this section, we introduce a general task model and a corresponding data model that form the foundation for multi-stage visual data selection algorithms.

\subsection{Task Model}
To describe the sensing objectives and constraints, a task is defined as a six-tuple:
\begin{equation}
	\text{task} = \{ \text{tid}, \text{type}, \text{whr}, \text{whn}, \text{angInter}, \text{altRange} \}.
\end{equation}
\begin{table}[!h]
    \centering
    \caption{Task Model Parameters}
    \label{tab:task-model}
    \begin{tabularx}{\columnwidth}{@{} >{\centering\arraybackslash\hsize=.6\hsize}X 
                                      >{\centering\arraybackslash\hsize=1.2\hsize}X 
                                      >{\centering\arraybackslash\hsize=1.2\hsize}X @{}}
        \toprule
        Symbol & Description & Example \\ \midrule
        $\text{tid}$ & Unique task ID & 1 \\
        $\text{type}$ & Accepted formats & (.jpg, .jpeg) \\
        $\text{whr}$ & Target GPS coordinates & (N34.246, E108.904) \\
        $\text{whn}$ & Time interval & (03141000, 03141800) \\
        $\text{angInter}$ & Angle granularity & $\pi/4$ \\
        $\text{altRange}$ & Altitude range & (0, 20m) \\
        \bottomrule
    \end{tabularx}
\end{table}

where \texttt{tid} denotes the unique task ID, \texttt{type} the expected image format, \texttt{whr} the target location, and \texttt{whn} the valid capture time range. The optional parameters \texttt{angInter} and \texttt{altRange} define viewpoint diversity constraints in angle and altitude, enabling multi-perspective coverage.

These task-level constraints help reduce redundant captures and promote viewpoint diversity by guiding participants to contribute complementary images across spatial and angular dimensions.

\subsection{Data Model}

Each image is described using a standardized data model as an eight-tuple:
\begin{equation}
	\text{data} = \{ \text{pid}, \text{tid}, \text{wid}, \text{type}, \text{time}, \text{locat}, \text{heig}, \text{resol} \}.
\end{equation}
\vspace{-0.3em}
\begin{table}[h]
    \centering
    \setlength{\tabcolsep}{2pt}
    \caption{Data Model Parameters}
    \label{tab:data-model}
    \begin{tabularx}{\columnwidth}{@{} >{\centering\arraybackslash\hsize=.6\hsize}X 
                                      >{\centering\arraybackslash\hsize=1.2\hsize}X 
                                      >{\centering\arraybackslash\hsize=1.2\hsize}X @{}}
        \toprule
        Symbol & Description & Example \\ \midrule
        $\text{pid}$ & Unique image ID & 101 \\
        $\text{tid}$ & Associated task ID & 1 \\
        $\text{wid}$ & Contributor ID & 5 \\
        $\text{type}$ & File format & .jpg \\
        $\text{time}$ & Capture time & 202403141530 \\
        $\text{locat}$ & GPS coordinates & (N34.246, E108.905) \\
        $\text{heig}$ & Altitude & 10.2m \\
        $\text{resol}$ & Resolution & 1080p \\ \bottomrule
    \end{tabularx}
\end{table}

where each parameter records key metadata useful for filtering and selection.

This model captures spatial, temporal, and quality attributes essential for metadata-based filtering. For instance, images with similar timestamps and locations but low resolution or redundant altitudes can be excluded early to improve transmission efficiency and downstream processing.

Together, the task and data models provide structured support for the multi-stage selection algorithm, enabling efficient, scalable filtering of heterogeneous visual crowdsensing data.
\begin{figure*}[t]
    \centering
    \includegraphics[width=0.95\textwidth]{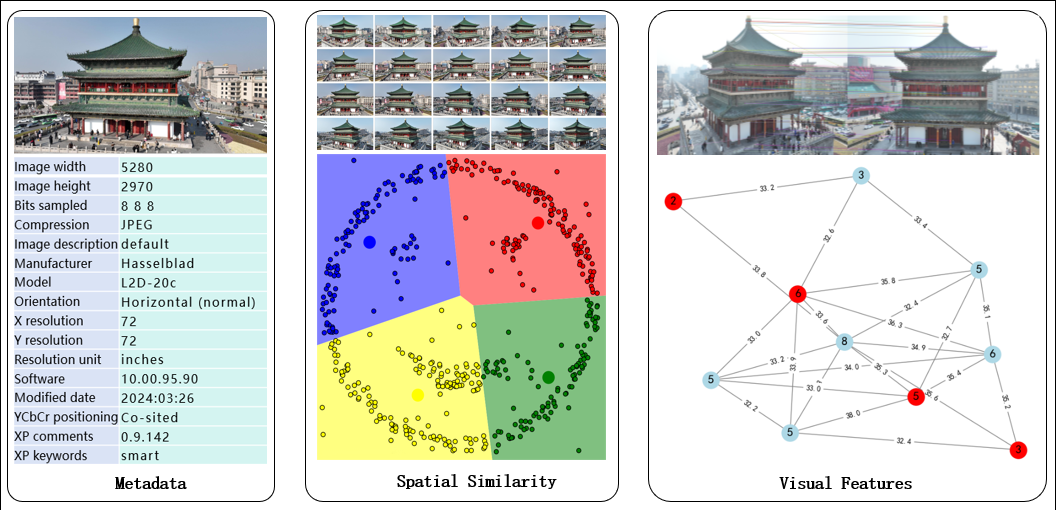}
    \caption{Overview of Multi-Stage Visual Data Selection Process}
    \label{fig:chapter3all}
\end{figure*}
\section{Multi-Stage Visual Data Selection Method}\label{chap3_4}

This section introduces \textbf{Tri-Select}, a three-stage visual data selection algorithm designed to reduce redundancy and improve representativeness in large-scale crowdsensing datasets. The method processes raw image data $P$ by sequentially applying metadata-based filtering, spatial clustering, and visual feature analysis, yielding a final subset $P_r$ of at most $B$ images. Figure~\ref{fig:chapter3all} illustrates the overall pipeline.

\subsection{Stage I: Metadata-Based Pre-Selection}\label{chap3_4_1}

This stage aims to efficiently filter out irrelevant or low-quality images using low-dimensional metadata, such as file format, timestamp, GPS coordinates, altitude, and resolution. It serves as a lightweight edge-side preprocessing step to reduce communication and computation costs in later stages.

Four parallel filters are applied:
\begin{itemize}
    \item \textbf{Format Filtering:} retain only images in allowed formats (e.g., JPEG, PNG);
    \item \textbf{Spatiotemporal Filtering:} keep images within the valid spatial radius and time window;
    \item \textbf{Altitude Filtering:} enforce altitude constraints to balance coverage from different perspectives;
    \item \textbf{Resolution Filtering:} remove low-quality images below a threshold (e.g., $360p$).
\end{itemize}

\begin{algorithm}[!h]
	\caption{Metadata-Based Pre-Selection}
	\label{alg:preselection}
	\begin{algorithmic}[1]
		\Require Dataset $P$, time range $[T_{\text{start}}, T_{\text{end}}]$, location radius $[D_{\min}, D_{\max}]$, altitude range $\text{altRange}$, resolution threshold $\text{resol}_{\min}$
		\Ensure Filtered subset $P_v$
		\State Initialize filter sets for format, time, GPS, altitude, resolution
		\For{each image $pid \in P$}
		    \If{format valid} \State add to $\text{format\_filtered}$ \EndIf
		    \If{time valid} \State add to $\text{time\_filtered}$ \EndIf
		    \If{GPS distance valid} \State add to $\text{gps\_filtered}$ \EndIf
		    \If{altitude valid} \State add to $\text{alt\_filtered}$ \EndIf
		    \If{resolution valid} \State add to $\text{quality\_filtered}$ \EndIf
		\EndFor
		\State $P_v \gets$ intersection of all filtered sets
		\State \Return $P_v$
	\end{algorithmic}
\end{algorithm}

This step reduces dataset size while preserving task-relevant images, making it ideal for edge-device deployment.

\subsection{Stage II: Spatial Similarity Clustering}\label{chap3_4_2}

After pre-filtering, the remaining data $P_v$ is grouped by spatial and directional similarity to organize images captured from similar perspectives. Spectral clustering is employed due to its robustness in handling non-convex clusters.

\subsubsection{Feature Vector Construction}

Each image is converted into a feature vector $\mathbf{f}_i$ combining relative location and shooting direction:
\begin{equation}
	\mathbf{f}_i = \left[ \frac{x_i - x_t}{\sigma_x}, \frac{y_i - y_t}{\sigma_y}, \cos\theta_i, \sin\theta_i \right]
\end{equation}
where $(x_t, y_t)$ is the target center, and $\theta_i$ is the capture angle. Normalization by $\sigma_x, \sigma_y$ ensures scale invariance.
\vspace*{-1in}
\newpage
\subsubsection{Similarity Matrix and Clustering}
Using the RBF kernel, the similarity between each pair of images is:
\begin{equation}
	S_{ij} = \exp\left(-\frac{\|\mathbf{f}_i - \mathbf{f}_j\|^2}{2\sigma^2}\right)
\end{equation}

Spectral clustering is then performed by computing the Laplacian matrix, extracting eigenvectors, and applying $k$-means in reduced space. The optimal cluster number $k$ is determined by silhouette score:
\begin{equation}
\text{Silhouette} = \frac{1}{N} \sum_{i=1}^{N} \frac{b(i) - a(i)}{\max\{a(i), b(i)\}}
\end{equation}

where $a(i)$ and $b(i)$ denote intra- and inter-cluster distances. This stage yields $N$ visually coherent clusters.

\subsection{Stage III: Visual Feature-Based Selection}\label{chap3_4_3}

The final stage selects a diverse and representative subset from the $N$ clusters using SIFT descriptors and graph-based optimization.

\subsubsection{SIFT Feature Extraction and Similarity}

For each image $I_i$, SIFT is used to extract keypoint descriptors $\mathbf{D}_i$. The similarity $S_{ij}$ between two images is computed using FLANN with Gaussian weighting:
\begin{equation}
S_{ij} = \frac{1}{N} \sum_{k=1}^{N} \exp\left(-\frac{\|d_{i,k} - d_{j,k}\|^2}{2\sigma^2}\right)
\end{equation}
\subsubsection{Graph Construction and MIS Search}

Construct a graph $G=(V, E)$ where each node is an image, and edges represent similarity above a threshold $\tau$:
\begin{equation}
w_{ij} =
\begin{cases}
S_{ij}, & \text{if } S_{ij} \geq \tau \\
0, & \text{otherwise}
\end{cases}
\end{equation}
We then search for a Maximum Independent Set (MIS), i.e., a subset of non-adjacent nodes, using a greedy strategy:
\begin{equation}
\max \sum_{v \in V} x_v \quad \text{s.t. } x_u + x_v \leq 1,\; \forall (u, v) \in E
\end{equation}

This stage ensures that the final output contains visually distinct and representative images, supporting high-quality crowdsensing analytics.
\begin{algorithm}[h]
	\caption{Visual Feature-Based Image Selection}
	\label{alg:image_selection}
	\begin{algorithmic}[1]
		\Require Clustered set $I = \{I_1, \dots, I_M\}$, similarity scale $\sigma$, target size $B$
		\Ensure Selected subset $I_r$
		\State Extract SIFT descriptors for all $I_i$
		\State Compute similarity matrix $S$ using FLANN
		\State Construct graph $G = (V, E)$ using threshold $\tau$
		\State Initialize $MIS \gets \emptyset$
		\While{$|MIS| < B$ and $G$ not empty}
		    \State Select node $v$ with lowest degree
		    \State Add $v$ to $MIS$, remove $v$ and neighbors from $G$
		\EndWhile
		\State \Return $I_r = \{ I_i \mid v_i \in MIS \}$
	\end{algorithmic}
\end{algorithm}
\section{Experiments and Results} \label{sec:experiment}
This section presents a comprehensive evaluation of the proposed multi-stage data selection algorithm. We first describe the datasets employed and the experimental setup, including implementation details and evaluation metrics. Then, we systematically report the experimental results for each individual stage, analyzing their respective impacts on the overall performance. Furthermore, we conduct comparative experiments against several state-of-the-art methods to demonstrate the superiority of the proposed approach in terms of data representativeness, redundancy reduction, and computational efficiency.

\subsection{Experimental Setup} \label{sec:experiment_setup}

\subsubsection{Datasets}

\begin{figure}[ht]
\centering
\begin{subfigure}[t]{0.47\columnwidth}
\centering
\includegraphics[width=0.83\linewidth]{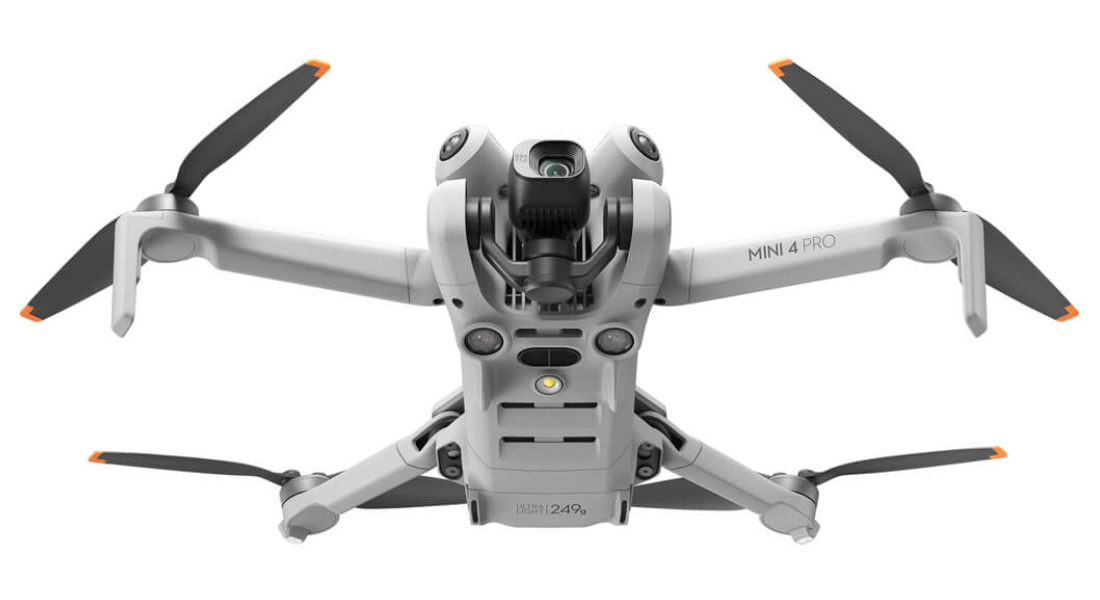}
\caption{DJI Mini 4 Pro}
\end{subfigure}
\hfill
\begin{subfigure}[t]{0.49\columnwidth}
\centering
\includegraphics[width=0.85\linewidth]{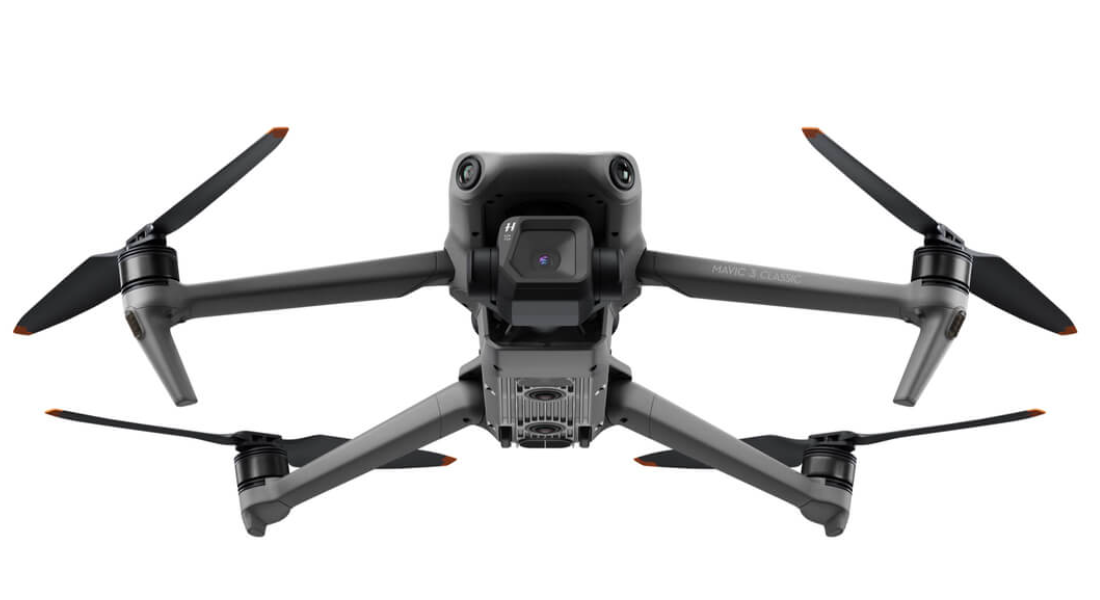}
\caption{DJI Mavic 3 Classic}
\end{subfigure}
\caption{UAVs used in data collection}
\label{fig:3DJI}
\end{figure}

We use both self-collected and public datasets. The self-collected dataset includes 348 images captured by 22 volunteers using drones and smartphones around a university campus. Each image is tagged with metadata including GPS coordinates, angles, altitude, timestamp, and resolution. Figure~\ref{fig:3DJI} shows the UAVs used. After preprocessing to remove low-quality samples, the dataset is divided into three subsets: NPU, TOWER, and NORMAL. Table~\ref{tab:dataset-info} summarizes key statistics.

\begin{table}[!h]
	\centering
	\caption{Image Dataset Summary}
	\label{tab:dataset-info}
    \begin{tabular}{cccc}
		\toprule
		Dataset & Images & Participants & Resolutions \\ \midrule
		NPU & 137 & 5 & $(5280\times2970)$, $(4032\times2268)$ \\
		TOWER & 264 & 10 & $(5280\times2970)$, $(3024\times3042)$ \\
		NORMAL & 185 & 7 & $(5280\times2970)$, $(960\times540)$ \\ \bottomrule
	\end{tabular}
\end{table}

To test generalizability of Stage III, we also use the COIL-100 dataset, which contains 7200 images of 100 objects taken from varying angles.

\subsubsection{Experimental Procedure}

Experiments were conducted independently for each stage:

- \textbf{Stage I:} Apply metadata constraints (format, time, location, altitude, resolution) to filter the dataset into a high-quality subset $P_v$.

- \textbf{Stage II:} Perform spectral clustering on $P_v$ using spatial and angular features to group similar viewpoints.

- \textbf{Stage III:} Use SIFT features and similarity graph-based MIS selection to extract $B$ representative images from each cluster.

\subsection{Results and Analysis}

\subsubsection{Stage I: Metadata-Based Pre-selection}

Table~\ref{tab:preselection-results} shows that our metadata pre-selection reduced image volume by 18.6\% to 29.2\%, with minimal loss of useful content. The average filtering time was 0.4s per 100 images, making it highly efficient for edge-side deployment.

\begin{table}[h]
	\centering
	\caption{Pre-selection Results}
	\label{tab:preselection-results}
	\begin{tabularx}{\columnwidth}{@{} >{\centering\arraybackslash}X 
                                      >{\centering\arraybackslash}X 
                                      >{\centering\arraybackslash}X 
                                      >{\centering\arraybackslash}X @{}}
		\toprule
		Dataset & Original & Selected & Reduction Rate \\ \midrule
		NPU & 137 & 97 & 29.2\% \\
		TOWER & 264 & 215 & 18.6\% \\
		NORMAL & 185 & 140 & 24.3\% \\ \bottomrule
	\end{tabularx}
\end{table}

\subsubsection{Stage II: Spectral Clustering}

\begin{figure}[!h]
    \centering
    \subfloat[Silhouette scores]{\includegraphics[width=0.5\linewidth]{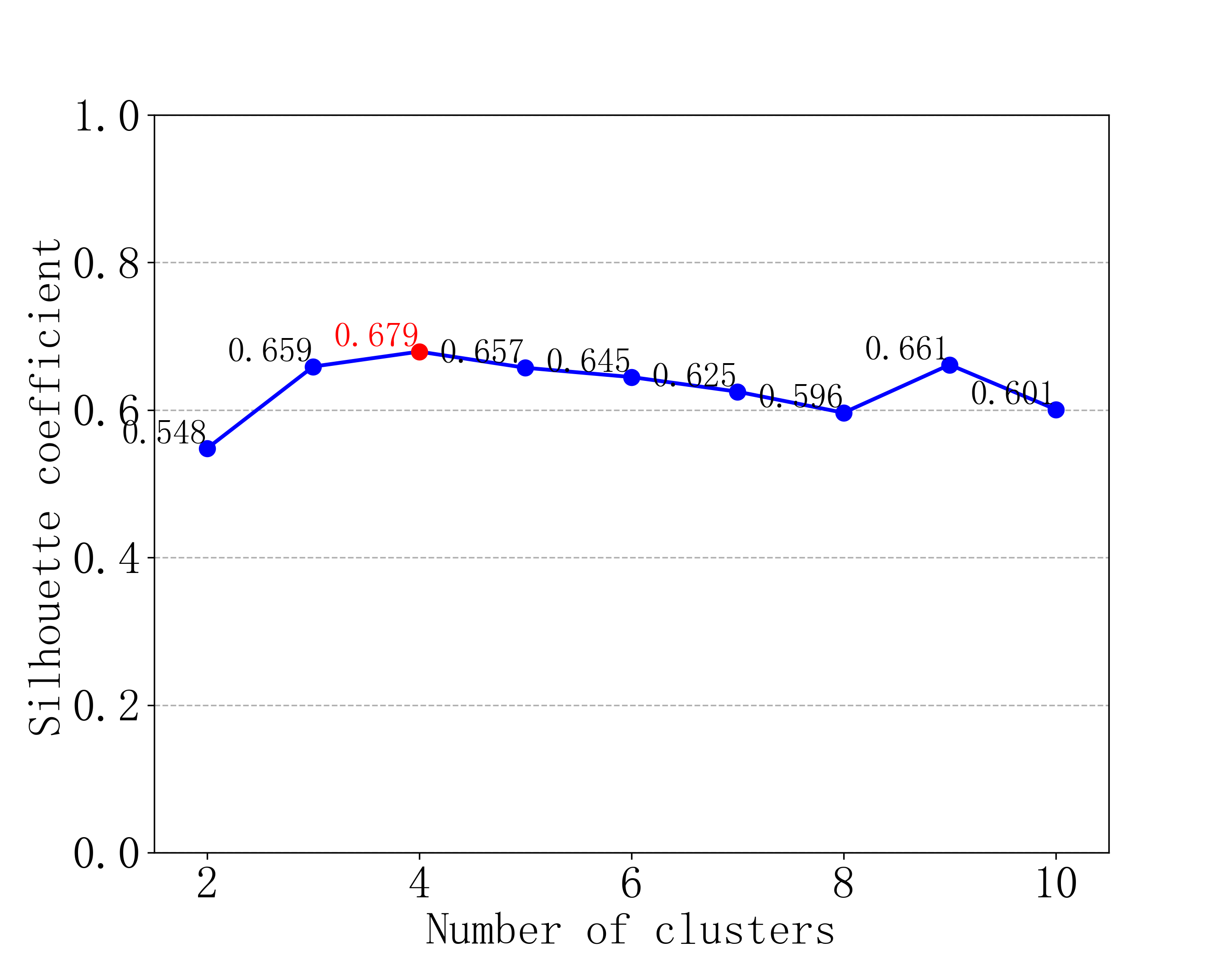}}
    \hfill
    \subfloat[Cluster scatter plot]{\includegraphics[width=0.5\linewidth]{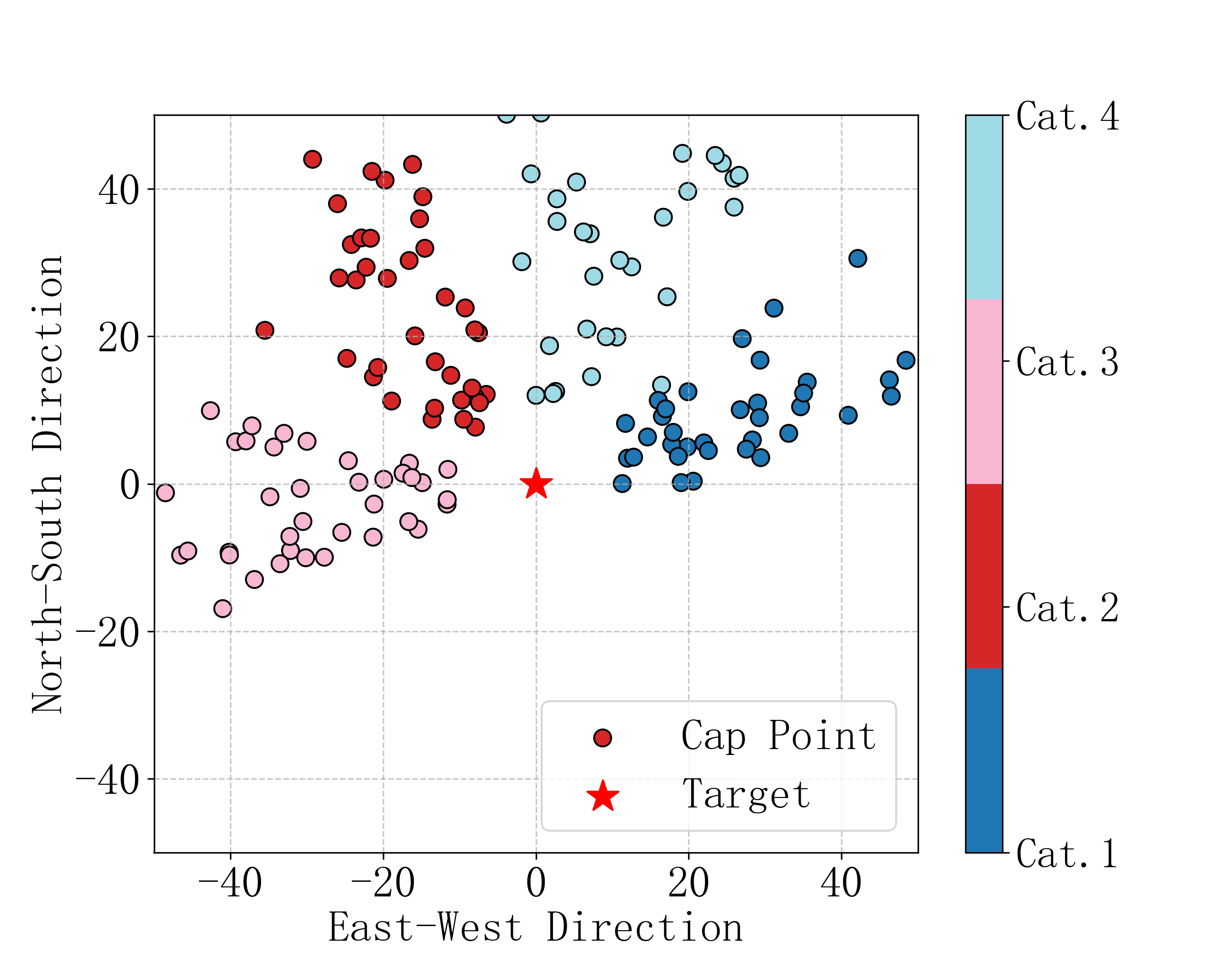}}
    \caption{Clustering results for NPU}
    \label{fig:cluster1}
\end{figure}

\begin{figure}[!h]
    \centering
    \subfloat[Silhouette scores]{\includegraphics[width=0.5\linewidth]{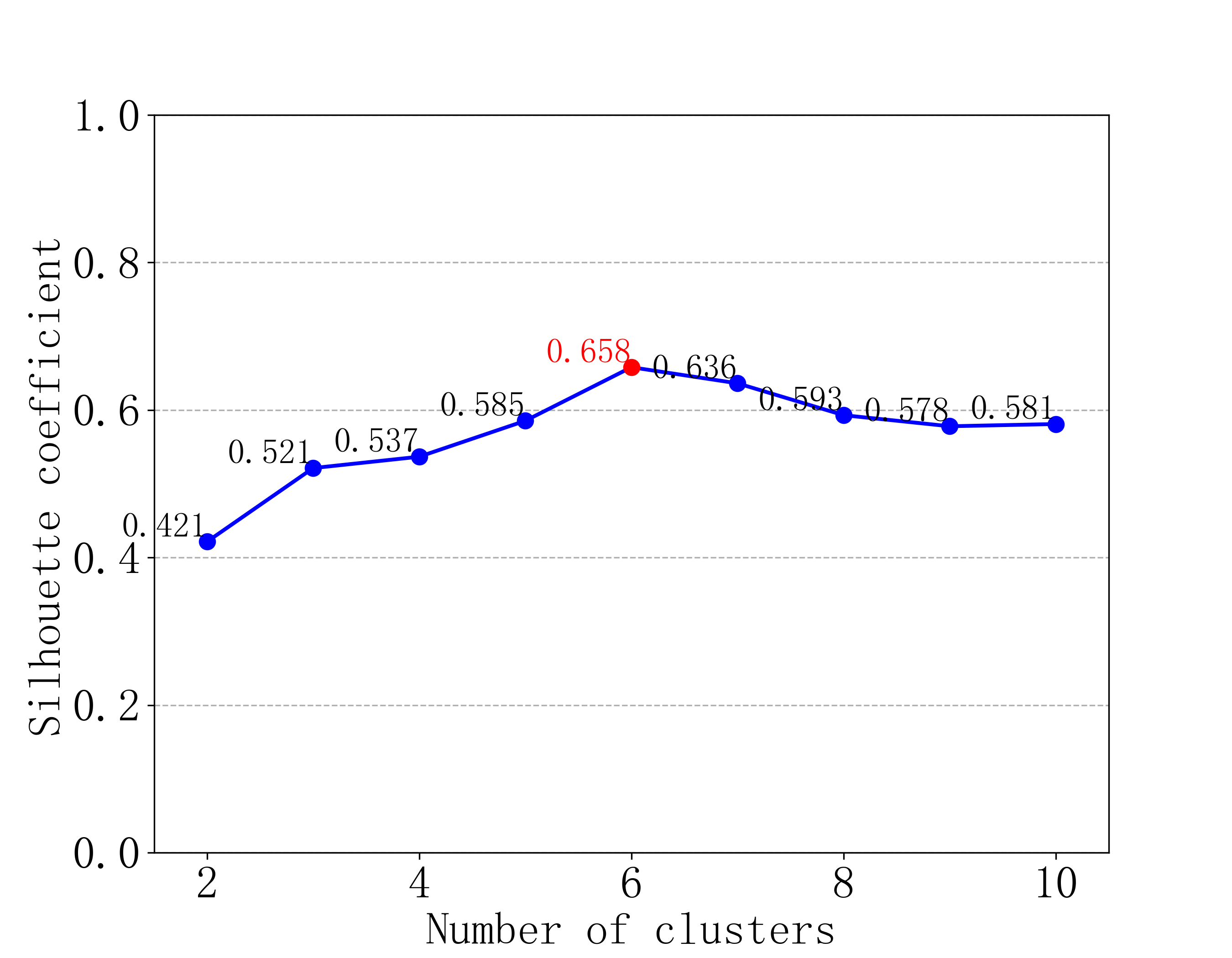}}
    \hfill
    \subfloat[Cluster scatter plot]{\includegraphics[width=0.5\linewidth]{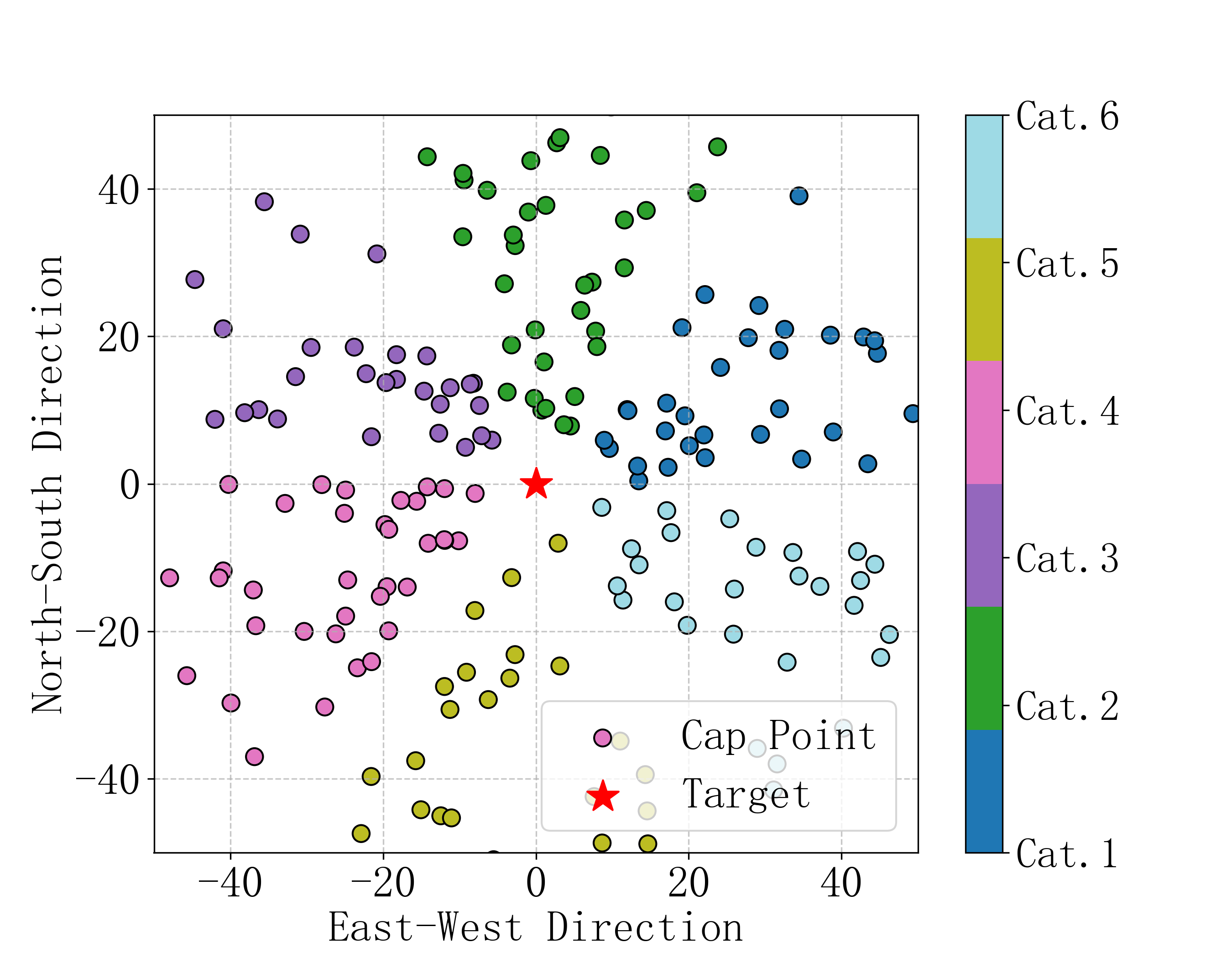}}
    \caption{Clustering results for TOWER}
    \label{fig:cluster2}
\end{figure}

\begin{figure}[!h]
    \centering
    \subfloat[Silhouette scores]{\includegraphics[width=0.5\linewidth]{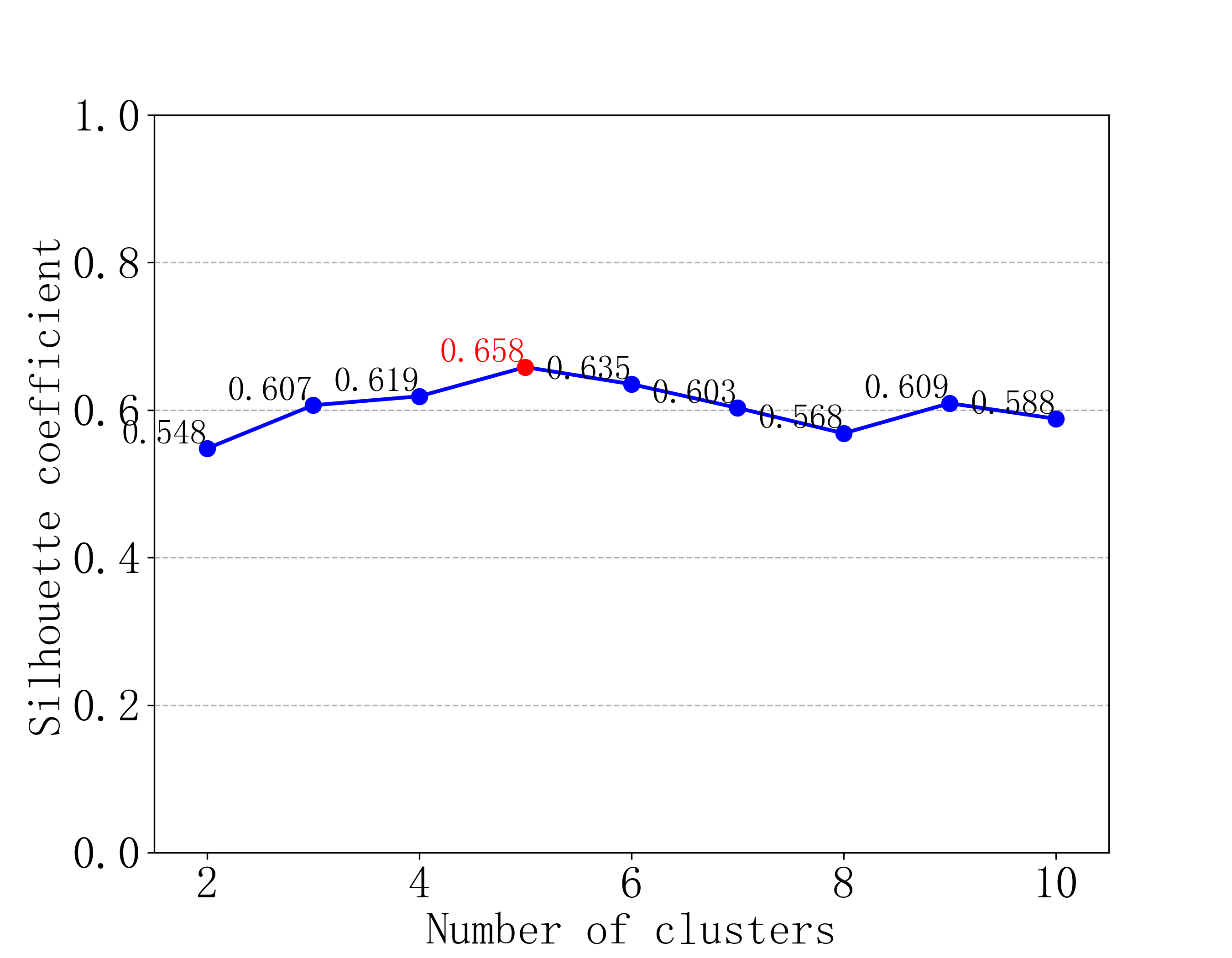}}
    \hfill
    \subfloat[Cluster scatter plot]{\includegraphics[width=0.5\linewidth]{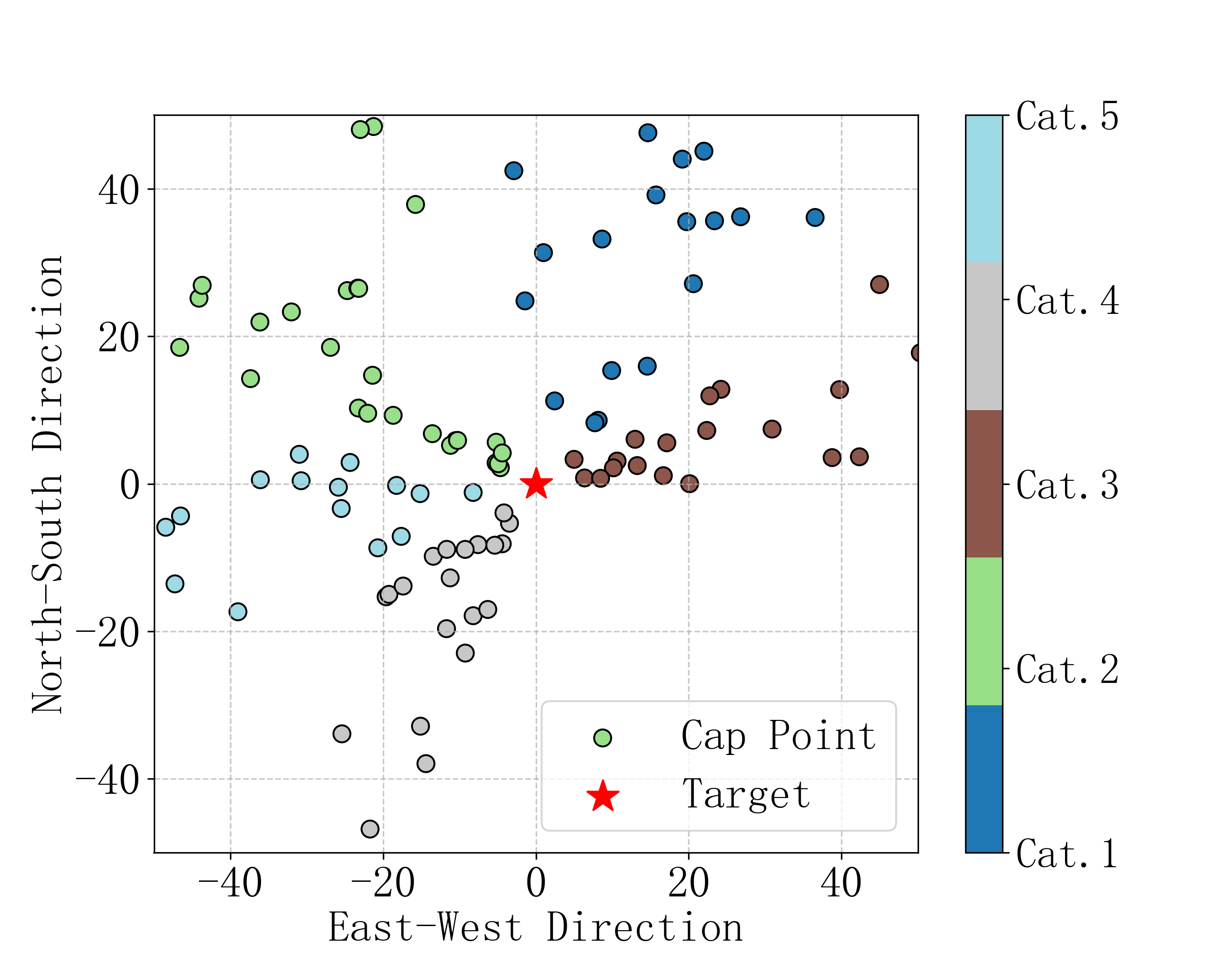}}
    \caption{Clustering results for NORMAL}
    \label{fig:cluster3}
\end{figure}

Figures~\ref{fig:cluster1} to~\ref{fig:cluster3} illustrate clustering results for three datasets. The optimal number of clusters $k$ is selected using the silhouette coefficient. For instance, $k=4$ yielded the best score (0.679) on NPU, while TOWER and NORMAL achieved optimal performance at $k=6$ and $k=5$, respectively. Each clustering output demonstrated strong spatial coherence and directional separation.

\subsubsection{Stage III: Visual Feature-Based Selection}

\begin{figure}[!h]
    \centering
    \includegraphics[width=1.0\linewidth]{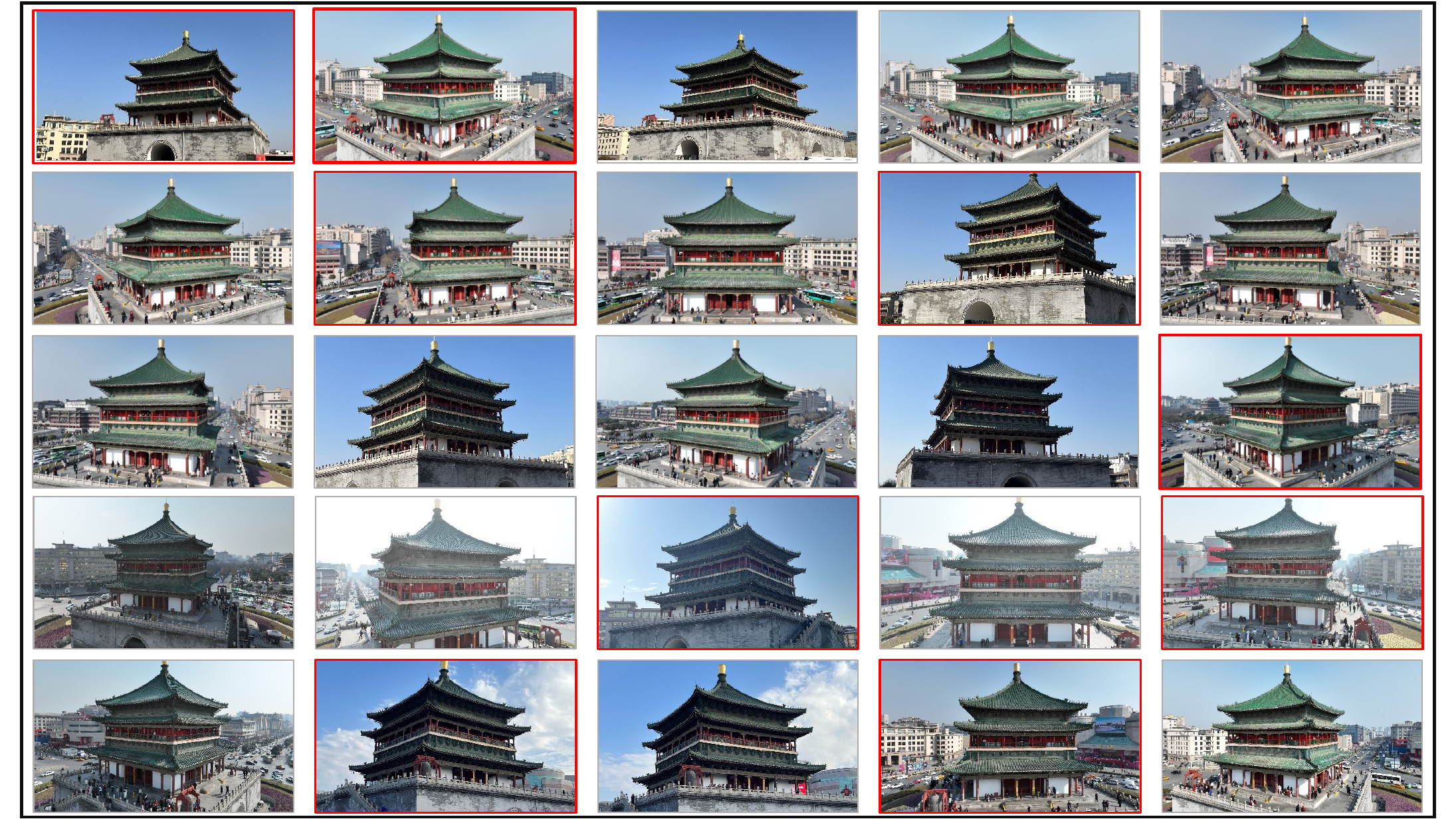}
    \caption{Selected images from TOWER dataset (highlighted in red)}
    \label{fig:stage3_tower}
\end{figure}

We applied the third-stage algorithm to both real-world and benchmark datasets to evaluate its ability to select diverse and representative images. For the TOWER dataset (Fig.~\ref{fig:stage3_tower}), the algorithm identified key viewpoints using SIFT and Maximum Independent Set (MIS) techniques, with red boxes marking final selections.

\begin{figure}[!h]
    \centering
    \includegraphics[width=1.0\linewidth]{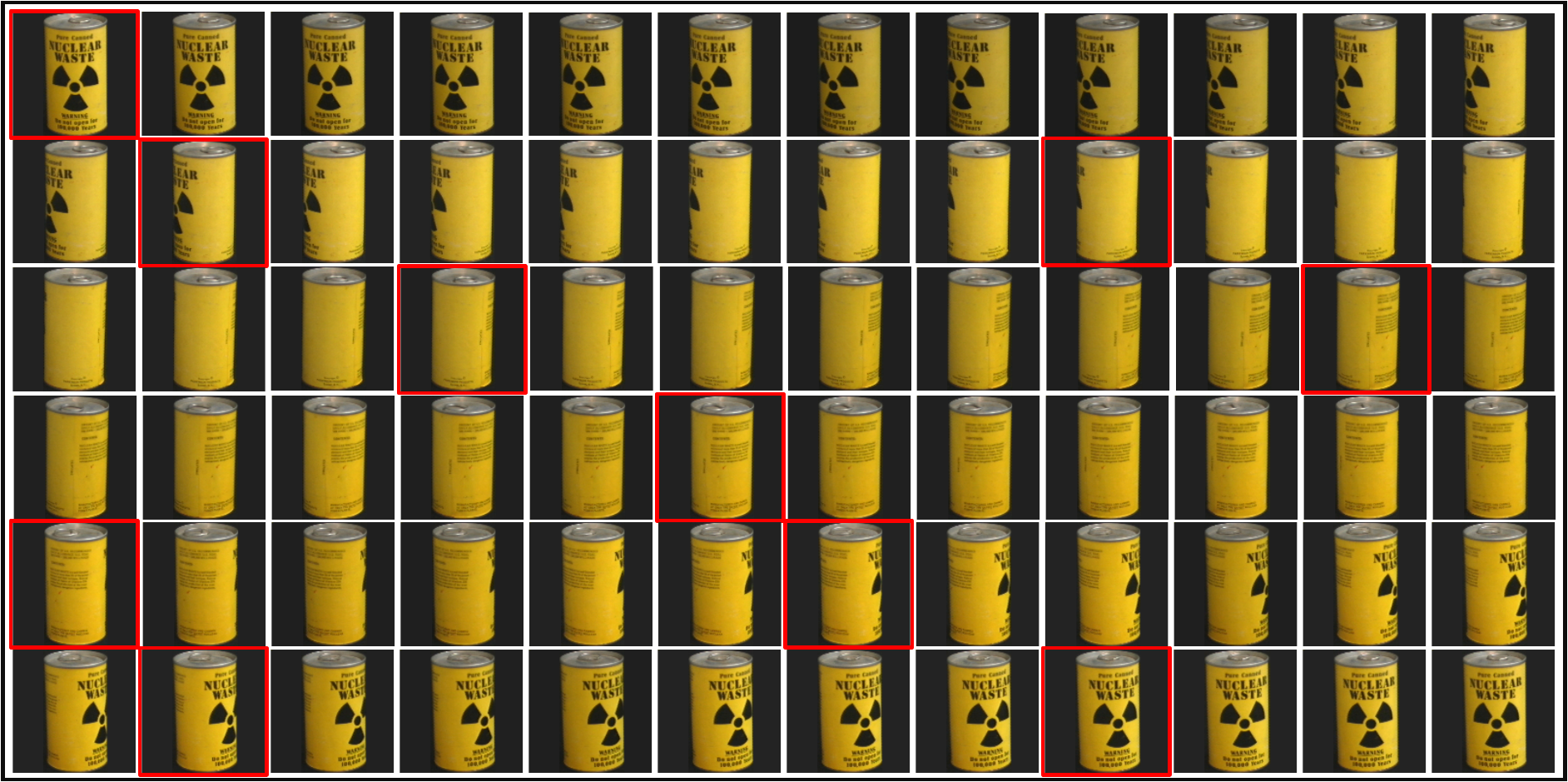}
    \caption{Selected images from COIL-100 dataset (highlighted in red)}
    \label{fig:stage3_before}
\end{figure}

On the COIL-100 dataset (Fig.~\ref{fig:stage3_before}), 10 representative images of object ID 66 were selected from 72 rotations. The algorithm preserved rotational diversity while avoiding redundant views. Total processing time was 20 seconds (0.2s per 100 images), confirming high scalability.

\subsubsection{Comparative Evaluation}

We compare our method (Ours) with three baselines: Ptree~\cite{guo2016picpick}, Hybrid~\cite{song2022hybrid}, and Cache~\cite{deng2023cache}. Evaluation metrics include selection rate and execution time.

\begin{figure}[!h]
    \centering
    \subfloat[Filtering rate]{\includegraphics[width=1.0\linewidth]{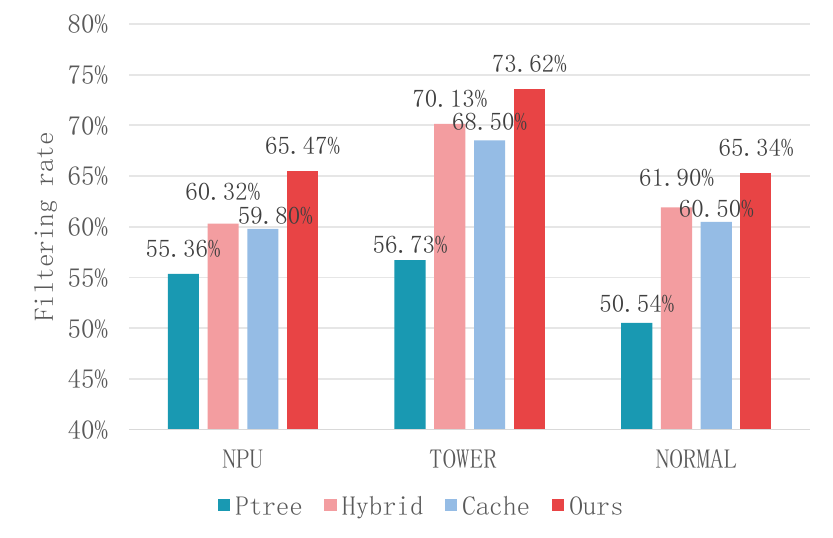}}
    \hfill
    \subfloat[Execution time]{\includegraphics[width=1.0\linewidth]{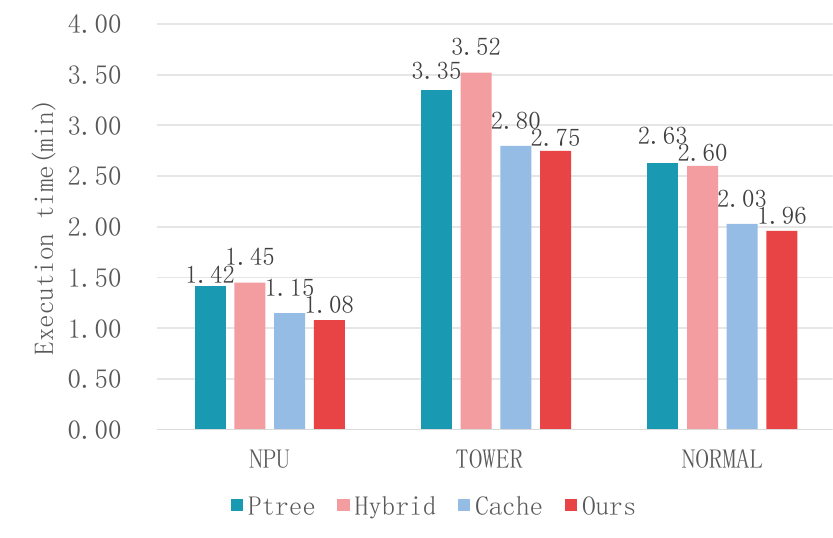}}
    \caption{Performance comparison across algorithms}
    \label{fig:exp3}
\end{figure}

As shown in Fig.~\ref{fig:exp3}, our method achieved the highest average selection rate (68.14\%) across three datasets, outperforming Hybrid (66.12\%), Cache (62.93\%), and Ptree (54.21\%). While Hybrid performed well on TOWER, it showed instability on other datasets. Cache balanced quality and speed, but lacked consistency. Our method maintained strong results due to its multi-stage pipeline.

In terms of execution time, our approach was the fastest (1.93 minutes average), ahead of Cache (1.99), Ptree (2.47), and Hybrid (2.52). Lightweight edge-compatible design and clustering efficiency enabled this improvement.

\subsubsection{Summary}

The experiments validate the effectiveness and scalability of the proposed algorithm. Key findings include:

\begin{itemize}
    \item \textbf{Stage I} reduced data volume by 24\% on average while retaining task-relevant content, with fast execution suitable for edge deployment.
    \item \textbf{Stage II} effectively grouped images using spatial and directional similarity. Clustering results were interpretable and stable across datasets.
    \item \textbf{Stage III} produced visually diverse, low-redundancy subsets using robust SIFT-based analysis and graph optimization.
\end{itemize}

Comparative studies demonstrated that the proposed method outperforms existing approaches in both selection quality and efficiency. Overall, our approach presents a scalable, efficient, and generalizable solution for high-quality data selection in large-scale visual crowdsensing.

\section{Conclusion}

This paper presents a multi-stage visual data selection algorithm for large-scale, multi-perspective crowdsensing tasks. The proposed method addresses redundancy, heterogeneity, and processing bottlenecks by sequentially applying metadata-based filtering, spatial similarity clustering, and visual-feature-guided selection. The three-stage pipeline effectively reduces data volume, organizes acquisition viewpoints, and identifies representative low-redundancy subsets.

Extensive experiments on real-world and benchmark datasets demonstrate superior performance over existing approaches in both selection quality and computational efficiency. The complementary roles of all three stages were validated through ablation studies. Future work will explore incorporating semantic features, task-specific constraints, and real-time adaptability to further enhance the method’s applicability in dynamic sensing scenarios.

\section{Acknowledgements}

The authors gratefully acknowledge the financial supports by different fundings. Kaixing Zhao is supported by the National Natural Science Foundation of China (No. 62407035 ), the China Postdoctoral Science Foundation (No. 2024M754226) and Shaanxi Province Postdoctoral Research Project Funding (No. W016308). Authors also thank the support of Hyper Creative Industry (Suzhou) Technology Co., Ltd.

\bibliographystyle{IEEEtran}
\bibliography{IEEEabrv,ref}

\end{document}